\documentclass[sigplan,screen]{acmart}

\AtBeginDocument{%
  }

\setcopyright{acmlicensed}
\copyrightyear{2024}
\acmYear{2024}
\acmDOI{XXXXXXX.XXXXXXX}

\renewcommand\footnotetextcopyrightpermission[1]{}
\usepackage{fancyhdr}

\usepackage{graphicx}
\usepackage{float}
\usepackage{soul, color, xcolor}
\acmSubmissionID{123-A56-BU3}
\usepackage{subcaption}
\usepackage{caption}
\usepackage{graphicx}
\usepackage{float}
\usepackage{svg}

\begin{document}
\pagestyle{empty}

\title{Real Face Video Animation Platform}

\author{Xiaokai Chen}
\email{kevincxk000@gmail.com}
\affiliation{%
  \institution{Harbin Institute of Technology}
  \city{Harbin}
  \country{China}
}

\author{Xuan Liu}
\email{cocopink1877@foxmail.com}
\affiliation{%
  \institution{Harbin Institute of Technology}
  \city{Harbin}
  \country{China}
}

\author{Donglin Di}
\authornote{Donglin Di is the corresponding author.}
\email{didonglin@lixiang.com}
\affiliation{%
  \institution{Space AI, Li Auto}
  \city{Beijing}
  \country{China}
}
\author{Yongjia Ma}
\email{maguire9993@gmail.com}
\affiliation{%
  \institution{Space AI, Li Auto}
  \city{Beijing}
  \country{China}
}

\author{Wei Chen}
\email{chenwei10@lixiang.com}
\affiliation{%
  \institution{Space AI, Li Auto}
  \city{Beijing}
  \country{China}
}

\author{Tonghua Su}
\email{thsu@hit.edu.cn}
\affiliation{%
  \institution{Harbin Institute of Technology}
  \city{Harbin}
  \country{China}
}
\renewcommand{\shortauthors}{Xiaokai Chen, Xuan Liu, Donglin Di, Yongjia Ma, Wei Chen, \& Tonghua Su}

\begin{abstract}
In recent years, facial video generation models have gained popularity. However, these models often lack expressive power when dealing with exaggerated anime-style faces due to the absence of high-quality anime-style face training sets. We propose a facial animation platform that enables real-time conversion from real human faces to cartoon-style faces, supporting multiple models. Built on the Gradio framework, our platform ensures excellent interactivity and user-friendliness. Users can input a real face video or image and select their desired cartoon style. The system will then automatically analyze facial features, execute necessary preprocessing, and invoke appropriate models to generate expressive anime-style faces. We employ a variety of models within our system to process the HDTF dataset, thereby creating an animated facial video dataset.
\end{abstract}



\begin{teaserfigure}
  \centering  
  \includegraphics[width=\textwidth]{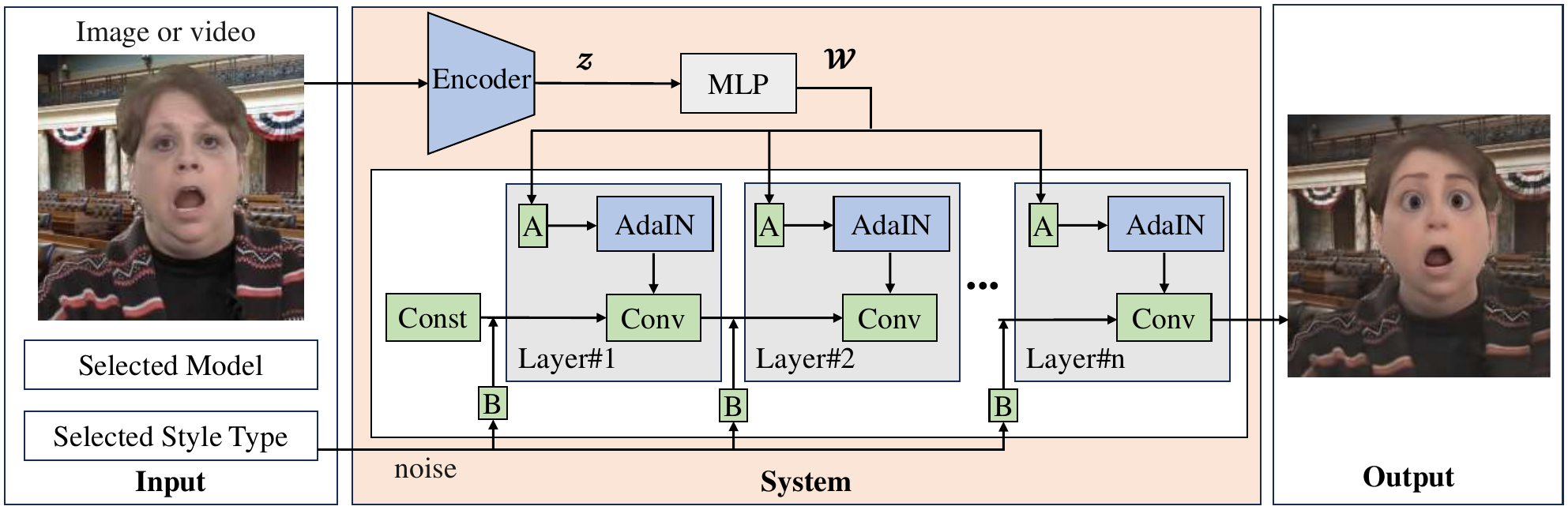}
  \caption{\textbf{Illustration of the video animation platform pipeline. ``AdaIN'' donates adaptive instance normalization. ``A'' donates a learned affine transform, and ``B'' applies learned per-channel scaling factors to the noise input.}}
  \Description{}
  \label{fig:teaser}
\end{teaserfigure}

\maketitle


\begin{figure*}[t]
	\centering
        \begin{subfigure}{0.23\linewidth}
		\centering
        \includegraphics[width=3.8cm,height=3.8cm]{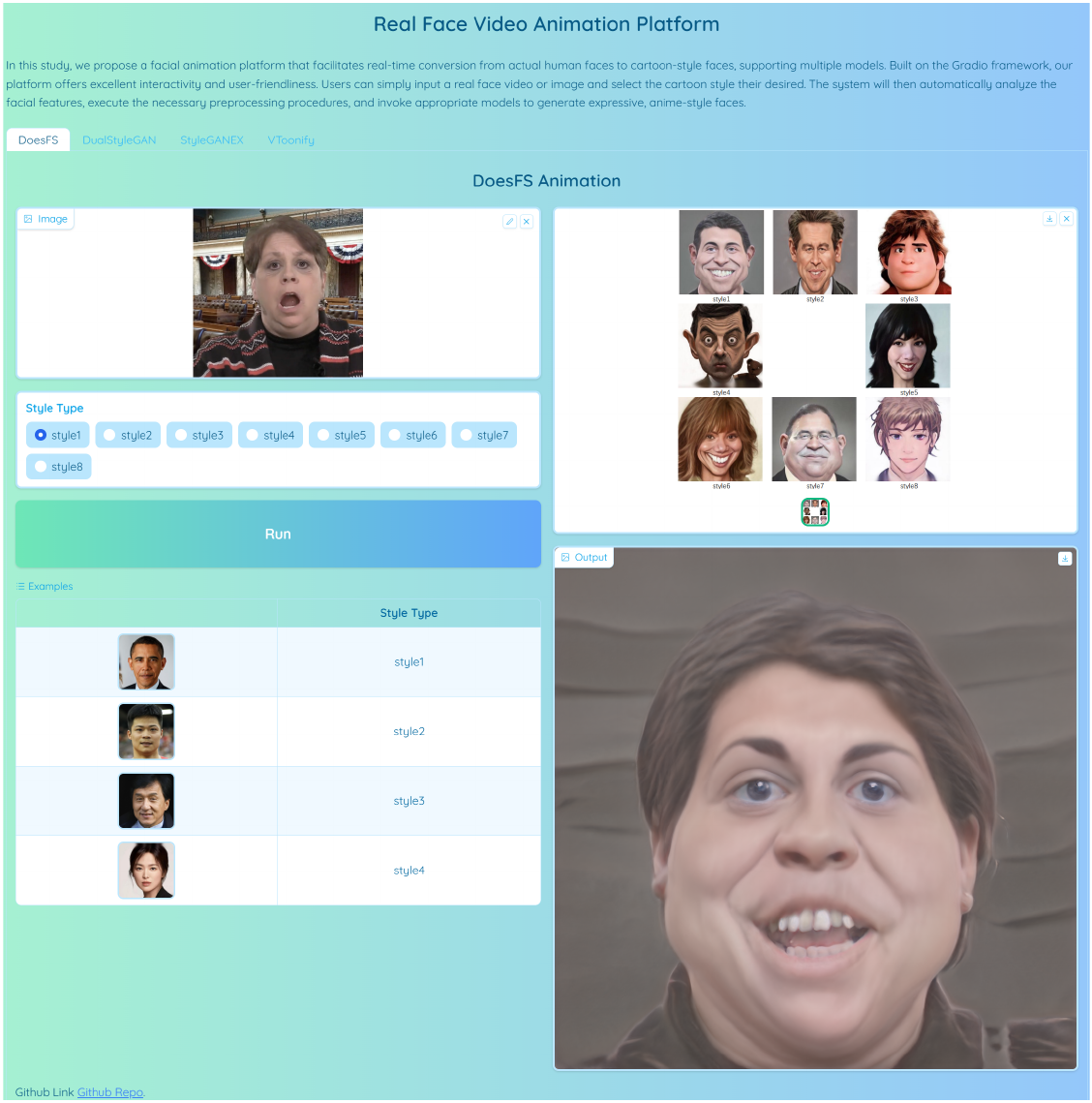}
		\caption{DoesFS}
		\label{DoesFS}
	\end{subfigure}
        \begin{subfigure}{0.23\linewidth}
		\centering
		\includegraphics[width=3.8cm,height=3.8cm]{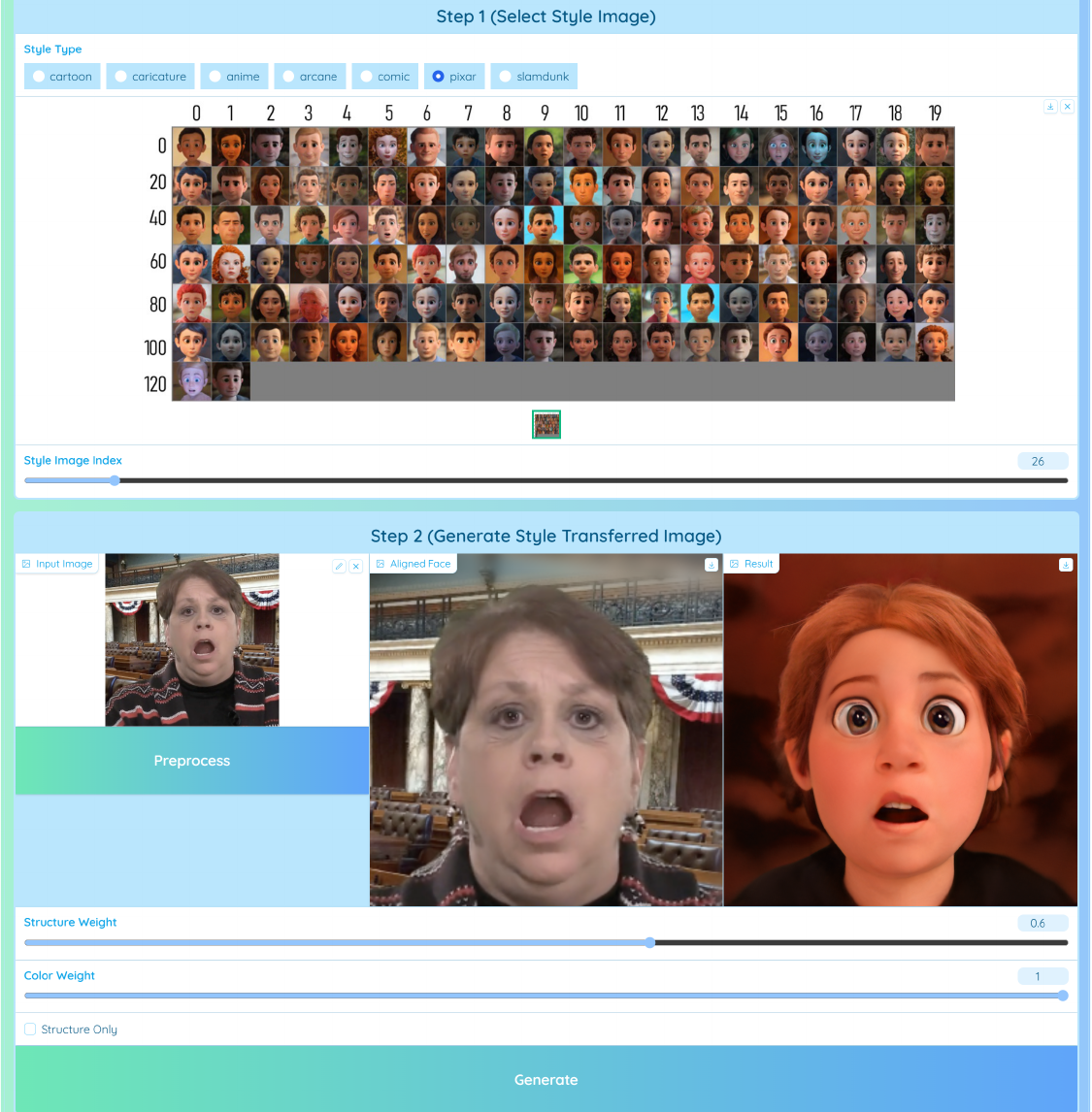}
		\caption{DualStyleGAN}
		\label{DualStyleGAN}
	\end{subfigure}
        \begin{subfigure}{0.23\linewidth}
		\centering
		\includegraphics[width=3.8cm,height=3.8cm]{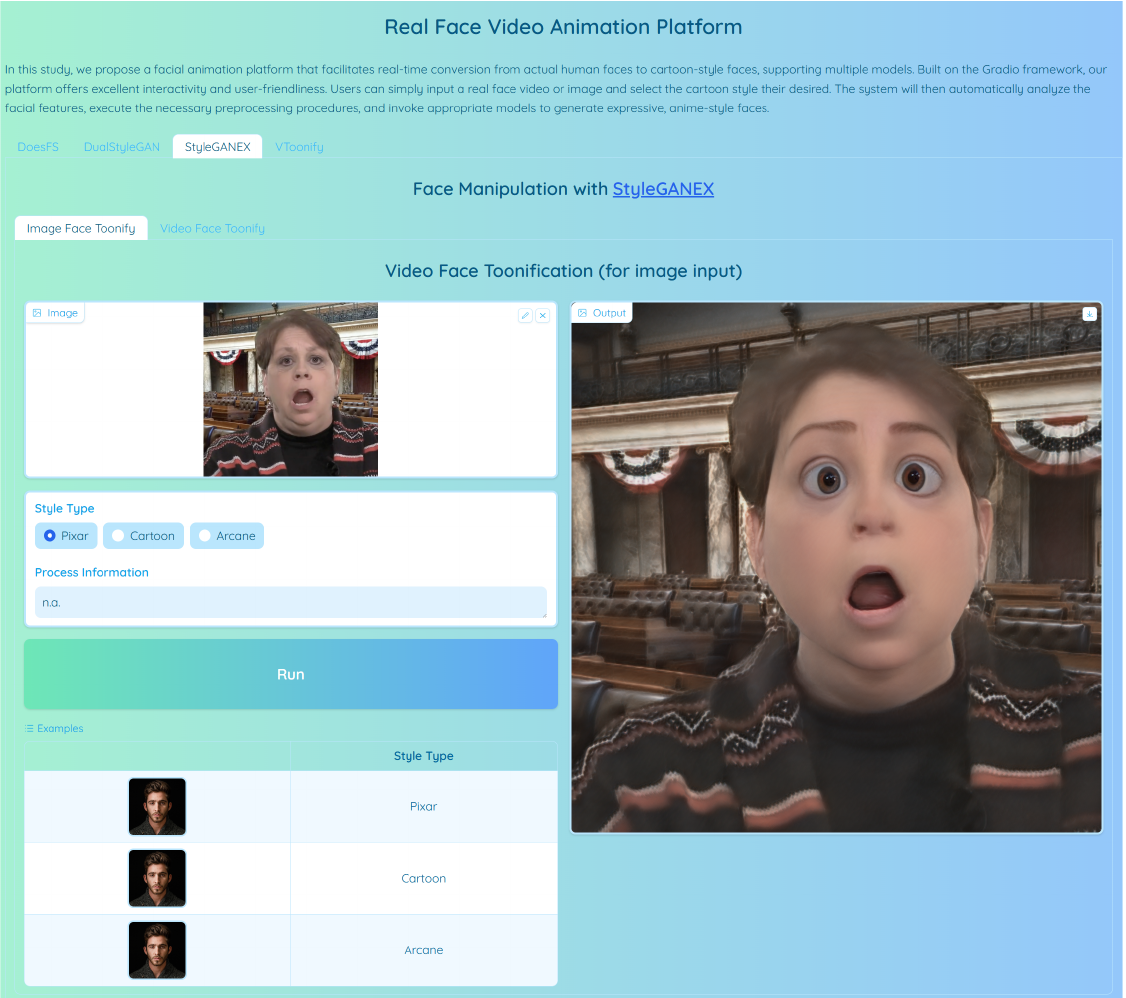}
		\caption{StyleGANEX}
		\label{StyleGANEX}
	\end{subfigure}
        \begin{subfigure}{0.23\linewidth}
		\centering
		\includegraphics[width=3.8cm,height=3.8cm]{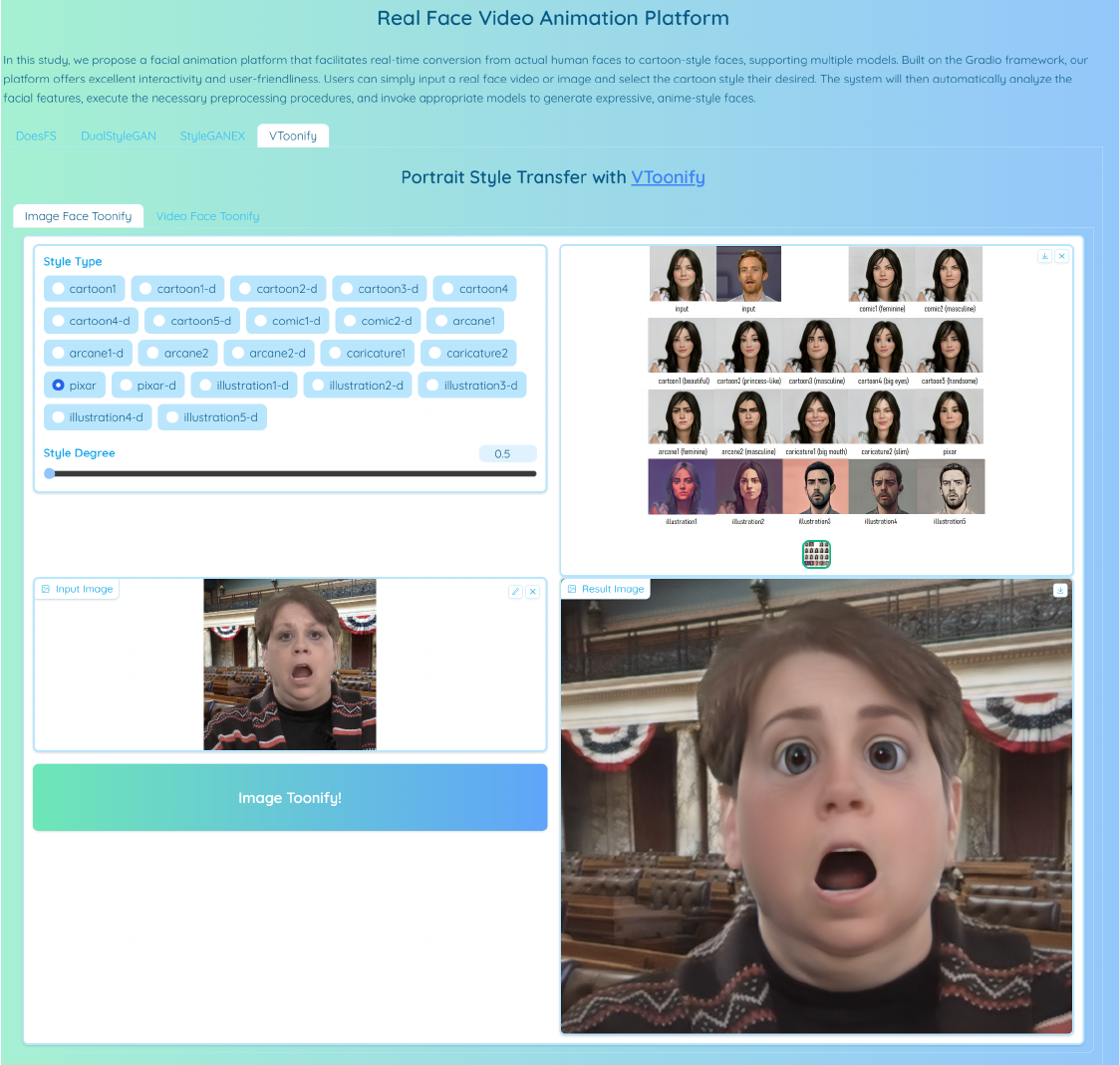}
		\caption{VToonify}
		\label{VToonify}
	\end{subfigure}
        \caption{\textbf{Screenshots of our demo interface.}}
        \label{Screenshot}
\end{figure*}

\section{Introduction}
Recently, anime-style portraits have gained prominence, frequently appearing in short videos, animations, and advertisements. The objective of portrait cartoonization algorithms is to transform real human face images into cartoon-style images, fundamentally a style transfer task. Deep learning methods such as neural style transfer (NST) \cite{Li_Chen_Yang_Wang_Lu_Yang_2017} and generative adversarial networks (GANs) \cite{Zhang_Lan_Yang_Hong_Wang_Yeo_Liu_Loy_2023} are commonly employed for this purpose. 

Significant advancements have been made in facial expression motion transfer using facial manipulation models \cite{animateanyone, emo, implicit}. However, these models frequently exhibit limitations in expressiveness when applied to anime-style faces. This shortcoming primarily stems from the lack of high-quality, anime-specific training datasets. The exaggerated expressions and unique facial feature configurations characteristic of anime faces pose significant challenges for models originally trained on real human faces. 

To obtain a high-quality cartoon-style facial dataset with rich expressiveness for fine-tuning existing models, we tested several facial stylization models on high-quality talking head video datasets such as HDTF \cite{Zhang_Li_Ding_Fan_2021}. By screening the results generated by various models, our system can help obtain a dataset of facial animations with a specific style that possesses enhanced expressive power. 

\section{Method}
The main objective of this work is to obtain a specific anime-style facial video dataset through a facial cartoonization model, which can assist in fine-tuning existing facial manipulation models. With the emergence of StyleGAN \cite{Karras_Laine_Aila_2019, Karras_Laine_Aittala_Hellsten_Lehtinen_Aila_2020}, GANs have become the mainstream choice for style transfer tasks due to their excellent image synthesis capabilities. The two model, StyleGANEX \cite{yang2023styleganex} and VToonify \cite{Yang_Jiang_Liu_Loy_2022} used in our demo are both advanced versions of StyleGAN. StyleGANEX employs dilated convolutions to rescale the receptive fields of StyleGAN's shallow layers, enhancing adaptability to variable-resolution facial images and improving accuracy in representing unaligned faces. VToonify leverages StyleGAN's high-resolution technology and the encoder's feature extraction capabilities to generate realistic yet aesthetically pleasing artistic portraits. Additionally, we experimented with various anime-style face generation models \cite{Yang_Jiang_Liu_Loy, zhou2024deformable, deformtoon3d, styletalk} . 

On our system, users can visually assess the generation results of various models for specific facial videos, filter out the desired data, and thereby produce cartoon facial videos with enhanced expressive power. 

\section{DEMO SYSTEM}

We have developed a web application based on the Gradio framework to facilitate users' access to face cartoonization. This application is accessible directly through a web browser. Figure \ref{fig:teaser} illustrates the framework of the proposed system, which can be divided into two main modules: customized anime style selection and generation result display. 

\textbf{Customized anime style selection}. Each model provided on the platform offers a wide range of style adjustment options, accompanied by sample displays of the corresponding styles to assist users in selecting their preferred anime style. Users can specify the desired output through buttons and sliders. 

\textbf{Generation result display}. After uploading an image or video and selecting the specific anime style, the system will invoke the selected model to generate the target image, using the chosen style as input noise for the network. The final result, meeting the user's expectations, will be displayed in the interface. 


Figure \ref{Screenshot} illustrates the user interface of the system. Users first select the model they wish to use and then upload the reference facial image or video. Subsequently, users can customize parameters such as style type and weight. Finally, the system invokes the model to generate and display the results on the interface. 

\section{CONCLUSION}
In this work, we propose a demo system that allows users to customize model and style selections for the anime-style transformation of facial images and videos. This platform enables users to upload their images or videos and compare the generation effects of various models. More importantly, the platform facilitates the collection of a high-quality anime-style facial dataset by stylizing real human faces to obtain diverse facial images and videos with varying stylistic nuances. It will enable more effective fine-tuning of existing models.



\bibliographystyle{unsrt}
\bibliography{ref}
\end{document}